\documentclass[11pt]{article}
\usepackage[]{ACL}

\usepackage{algorithm}
\usepackage{algorithmicx}
\usepackage{algpseudocode} 
\usepackage{subcaption}
\usepackage{makecell}
\usepackage{amssymb}
\usepackage[table]{xcolor}
\usepackage{amsmath}
\usepackage{booktabs}
\usepackage{multirow}
\usepackage{placeins}
\usepackage{array}
\usepackage{caption}
\usepackage{times}
\usepackage{latexsym}
\usepackage{pifont}
\usepackage[most]{tcolorbox}
\usepackage{xcolor}
\usepackage[T1]{fontenc}
\usepackage[utf8]{inputenc}
\usepackage{microtype}
\usepackage{inconsolata}
\usepackage{graphicx}

\title{Guideline Forest: Retrieval-Augmented Reasoning with Branching Experience-Induced Guidelines}

\author{
Jiaxiang Chen$^{1,*}$ \quad
Zhuo Wang$^{1,4*}$ \quad
Mingxi Zou$^{1}$ \quad
Qifan Wang$^{2}$ \quad
Zenglin Xu$^{1,3}$ \\
\\
$^{1}$Fudan University \quad
$^{2}$Meta \quad
$^{3}$SAIS \quad
$^{4}$SII
}

\usepackage{float}

\begin{document}
\maketitle

\begingroup
\renewcommand\thefootnote{}
\footnotetext{*Equal contribution.}
\addtocounter{footnote}{-1}
\endgroup

\begin{abstract}
Retrieval-augmented generation (RAG) has been widely adopted to ground large language models (LLMs) in external knowledge, yet it remains largely underexplored for improving reasoning. Existing methods either rely on online exploration during inference or heuristic supervision over reasoning trajectories, but they fail to effectively accumulate and reuse past reasoning experience.
We propose Guideline Forest, a retrieval-augmented reasoning framework that explicitly leverages experience to guide multi-step reasoning. The framework stores high-quality, label-consistent reasoning traces as reusable memory, retrieves relevant experiences for new problems, and induces them into structured guidelines that steer reasoning and enable controlled branching and aggregation.
Experiments on mathematical (GSM8K, MATH-500) and programming (MBPP, HumanEval) benchmarks demonstrate consistent improvements over strong reasoning baselines, including CoT, ReAct, ToT, FoT, and AFlow. Further analyses show that experience retrieval, guideline-induced diversity, and stepwise aggregation are key to the framework’s effectiveness.
Beyond single-model reasoning, Guideline Forest generalizes to enhance diverse reasoning paradigms and supports multi-model collaboration, highlighting its flexibility and scalability.
\end{abstract}

\section{Introduction}

\begin{figure}[!t]
    \centering
    \includegraphics[width=0.47\textwidth]{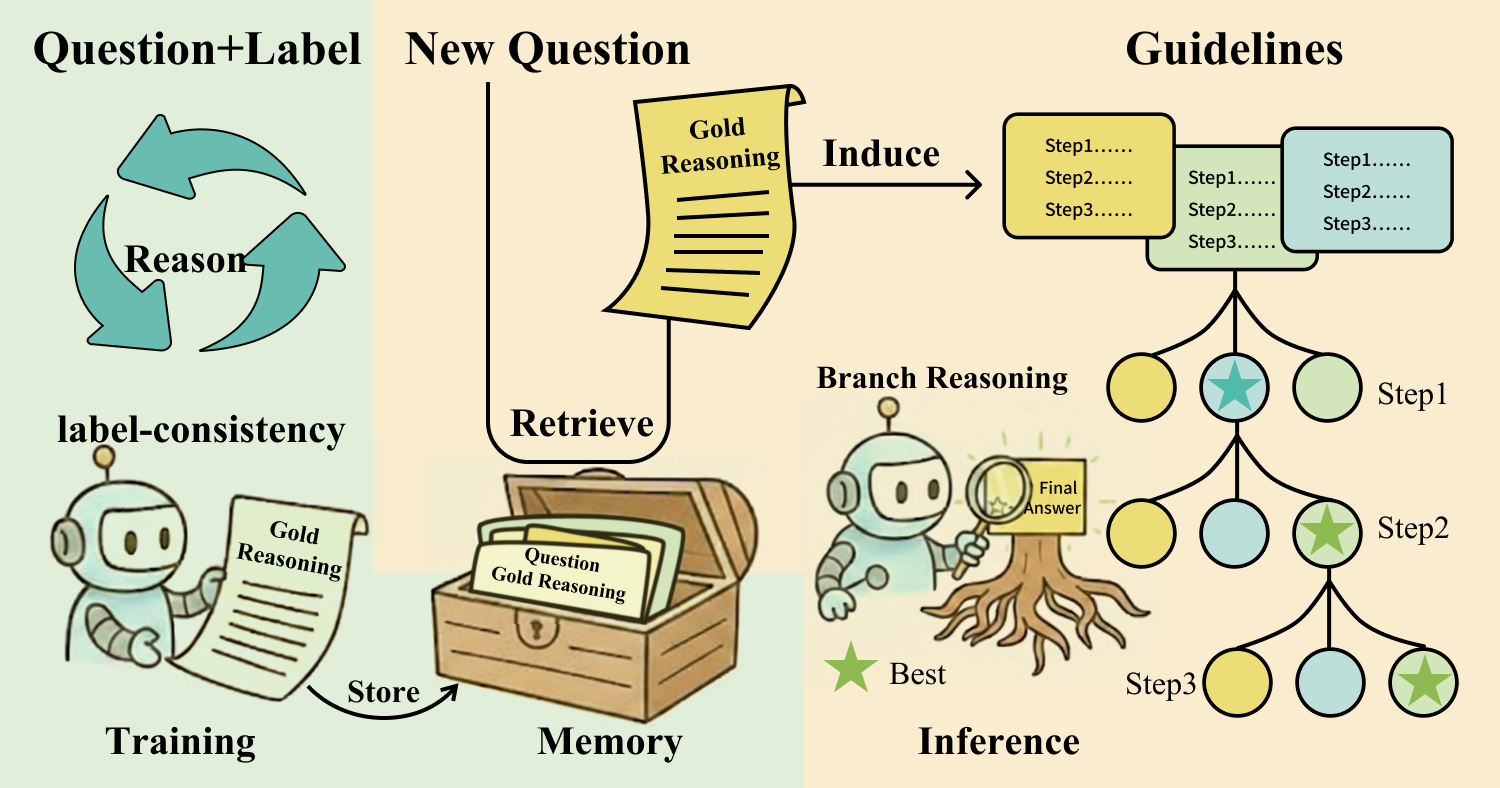}
    \caption{A conceptual illustration of our approach.
The agent is trained with question–label pairs to produce high-quality, label-consistent \textbf{gold reasoning} stored in memory.
When facing a new question, it retrieves relevant gold reasoning, induces \textbf{guidelines}, and solves the problem through stepwise branch reasoning.}
    \label{fig:illustration}
    \vspace{-0.05cm}
\end{figure}

A key characteristic of human reasoning is its flexibility and reliance on accumulated experience.
When solving complex problems, humans rarely commit to a single fixed line of thought.
Instead, they recall relevant examples, reuse effective strategies, and explore multiple reasoning paths.
These paths are iteratively compared, revised, and integrated through reflection and self-correction, before converging on a coherent solution.
Such non-linear reasoning enables new problems to be grounded in prior experience, yielding robust and adaptable problem solving.

In contrast, large language models (LLMs)~\cite{gpt4,llama3.1,deepseekv3,qwen2.5} typically generate reasoning as a single linear sequence, without explicit mechanisms to store, retrieve, or reuse prior reasoning experience.
Existing approaches attempt to mitigate this limitation in two main ways.
Exploration-based methods (e.g., CoT~\cite{cot,sc_cot}, ToT~\cite{tot}, FoT~\cite{fot}) generate multiple reasoning paths during inference, but rely on unguided search without leveraging accumulated experience.
Workflow-oriented methods (e.g., AFlow~\cite{zhang2024aflow}) learn preferred reasoning procedures under supervision, yet often lack adaptability when transferred to new problems.
As a result, prior reasoning is rarely reused in a systematic and structured manner.

Motivated by this gap, we propose \textbf{Guideline Forest}, a retrieval-augmented reasoning framework that grounds inference in accumulated reasoning experience.
Unlike conventional RAG~\cite{graphrag,lightrag,slimrag}, which retrieves external knowledge, Guideline Forest retrieves and reuses \emph{reasoning experience}.
During training, we construct a memory of \emph{label-consistent gold reasoning trajectories}.
Specifically, reasoning is first generated using standard chain-of-thought prompting; when the resulting answer is incorrect, we apply \emph{label-guided generation} to regenerate reasoning that is explicitly constrained to produce the correct label.
This process yields verified, coherent reasoning traces that are stored as reusable experience.

At inference time, given a new problem, the framework retrieves relevant gold reasoning trajectories and abstracts them into structured \emph{guidelines}.
These guidelines drive \emph{branch reasoning}, where multiple reasoning variants are expanded and refined in parallel.
A stepwise aggregation mechanism compares candidate branches at each stage and selects the most promising continuation, leading to more stable and consistent multi-step reasoning.
The same aggregation principle naturally extends to multi-model settings, where outputs from different models are jointly evaluated to mitigate individual biases and improve robustness.

We evaluate \textbf{Guideline Forest} on four benchmarks covering math and code reasoning (GSM8K, MATH-500, MBPP, HumanEval).
Across all settings, our framework consistently outperforms strong baselines including CoT, ReAct, ToT, FoT, and AFlow, while achieving a favorable accuracy--cost trade-off through experience-induced branching and step-wise aggregation.
Moreover, the induced guidelines generalize beyond our framework, improving existing reasoning methods and enabling robust multi-model collaboration.
Ablation studies further validate the roles of retrieval scale, path diversity, stepwise refinement, and aggregation.

\textbf{Our key contributions are as follows:}
\begin{itemize}
    \item We propose \textbf{Guideline Forest}, a retrieval-augmented reasoning framework that treats verified reasoning trajectories as reusable experience, extending RAG from knowledge retrieval to \emph{reasoning retrieval}.
    \item We introduce \textbf{label-guided generation} to construct label-consistent gold reasoning, together with a structured \textbf{branch reasoning and aggregation mechanism} for robust and coherent multi-step inference.
    \item Through extensive experiments on math and code benchmarks, we show that \textbf{Guideline Forest} consistently outperforms strong baselines, achieves a favorable \textbf{accuracy--cost trade-off}, and naturally supports effective multi-model collaboration; ablation studies further validate the individual contributions of key components within the framework.
\end{itemize}

\begin{figure*}[!t]
    \centering
    \includegraphics[width=\textwidth]{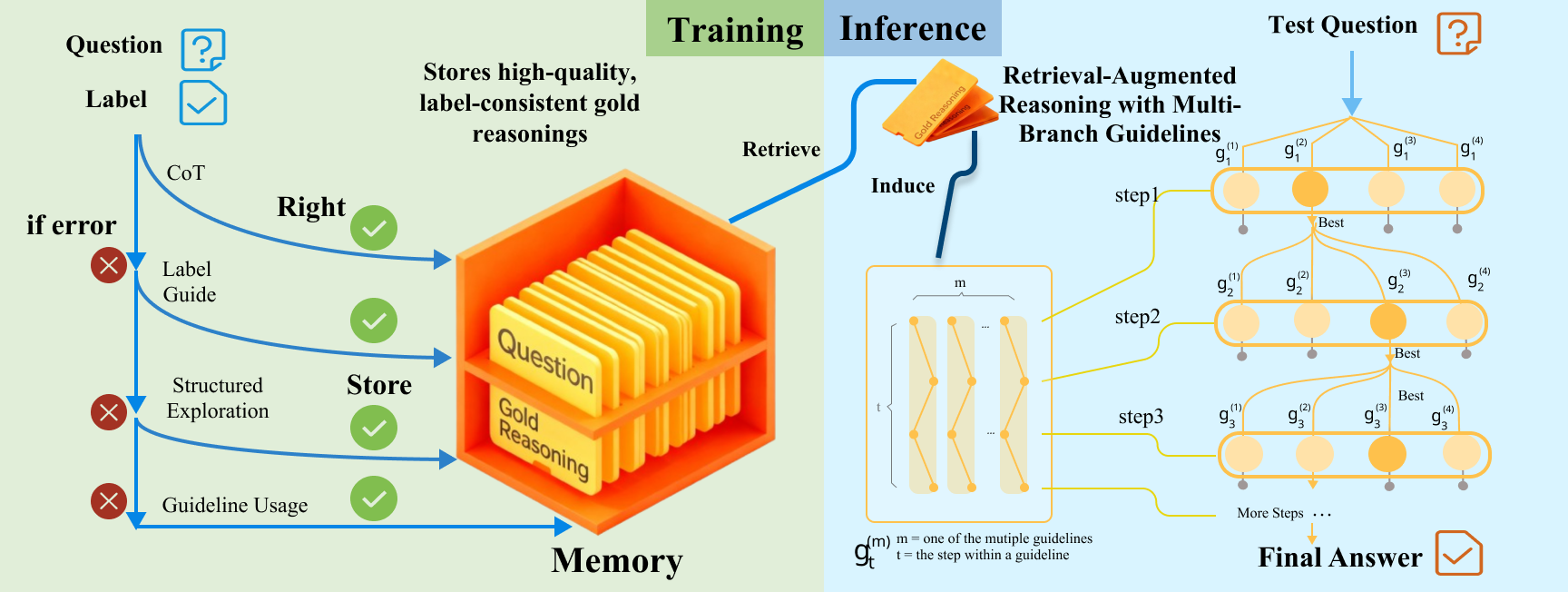}
    \caption{
\textbf{Overall framework of Guideline Forest.}
(\textbf{Left}) During training, the model learns to produce high-quality, label-consistent \emph{gold reasonings} through iterative correction—
first via chain-of-thought (CoT), then with label guidance, structured exploration (ToT), and memory-based guideline usage when necessary.
Verified trajectories are stored in a memory repository, forming a growing collection of reusable reasoning experience.
(\textbf{Right}) During inference, the system retrieves relevant gold reasonings, induces \emph{multi-branch guidelines}, and executes them in parallel, selecting the best at each step to refine reasoning and yield a higher-quality answer.
}
    \label{fig:framework}
\end{figure*}

\section{Related Work}

\subsection{Exploratory Reasoning}

Exploratory prompting methods improve LLM reasoning by sampling multiple reasoning paths at inference time~\cite{cot,got,tot,fot,react,selfrefine}. 
CoT~\cite{cot} generates explicit intermediate steps and has been widely applied to mathematical and programming tasks~\cite{mathprompter,reasonsurvey,code_understand,codesurvey}. 
ToT~\cite{tot} and FoT~\cite{fot} extend CoT with tree- or forest-structured exploration, while ReAct~\cite{react} interleaves reasoning with external actions.
However, these approaches construct reasoning trajectories from scratch for each instance and rely entirely on test-time exploration. 
They do not reuse prior reasoning experience, leading to inefficiency and limited generalization. 
Our work instead grounds reasoning in reusable trajectories accumulated during training.

\subsection{Search-based Reasoning}

Search-based methods cast reasoning as a structured search or workflow optimization problem~\cite{beats,hiar-icl,booststep,adas,zhang2024aflow}. 
AFlow~\cite{zhang2024aflow} and HaIR-ICL~\cite{hiar-icl} apply MCTS~\cite{mcts,mctssurvey} to explore reasoning operators or strategy spaces, while WorkflowLLM~\cite{fan2024workflowllm} focuses on workflow orchestration. 
BoostStep~\cite{booststep} retrieves reasoning steps from a case library, and AdaS~\cite{adas} allocates computation adaptively.
Although effective, these methods often incur high computational cost and depend on heuristic search or handcrafted operators. 
In contrast, our approach learns experience-grounded reasoning guidelines from verified examples and aggregates them efficiently during inference.

\subsection{Collaborative Inference}

Collaborative inference is closely related to ensemble learning, as both improve robustness by aggregating multiple predictors or experts at inference time~\cite{decisiontree,svm,classificationtree,Randomforests,adaboost,baggingboost,adaptivemoe,deepseekmoe}.
In reasoning, this idea appears in methods such as Self-Consistency~\cite{sc_cot}, Tree-of-Thought~\cite{tot}, and Forest-of-Thought~\cite{fot}, which aggregate multiple reasoning outputs during inference, as well as in multi-LLM systems like Avengers~\cite{avengers} and SMACS~\cite{Open-SourceLLMs} that leverage complementary model strengths.
Different from these approaches, our method performs collaborative inference at the level of reasoning paths, guided by reusable reasoning experience rather than pure test-time sampling, enabling structured branching and aggregation across paths and models.
This perspective frames our approach as a form of \textbf{branch-enhanced learning} driven by accumulated reasoning trajectories.

\section{Method}

\subsection{Preliminaries}

\paragraph{Multi-Step Reasoning Formulation.}
We study multi-step reasoning in large language models (LLMs) using latent reasoning trajectories.
Given an input problem $x$ and final answer $y$, the reasoning process is modeled as a sequence of intermediate steps
$\mathbf{z} = (z_1, \dots, z_T)$.
The prediction is formulated as a marginalization over latent trajectories:
\begin{equation}
P_\theta(y \mid x) = \sum_{\mathbf{z}} P_\theta(y \mid \mathbf{z}, x)\, P_\theta(\mathbf{z} \mid x),
\end{equation}
where $P_\theta(\mathbf{z} \mid x)$ represents trajectory generation and
$P_\theta(y \mid \mathbf{z}, x)$ predicts the answer conditioned on the full reasoning.
Existing methods such as CoT and ToT instantiate this formulation by sampling or searching trajectories at test time.

\paragraph{Trajectories vs. Guidelines.}
We distinguish between \emph{reasoning trajectories} and \emph{guidelines}.
A reasoning trajectory $\mathbf{z}$ is a concrete, instance-specific sequence of inference steps.
In contrast, a guideline $\mathbf{g} = (g_1,\dots,g_T)$ is an abstract, reusable plan induced from verified trajectories.
Guidelines capture general reasoning structure rather than exact solutions, and act as instructional scaffolds that guide step-wise generation.

\subsection{Training: Constructing Gold Reasoning}

Our training goal is to build a repository of verified \emph{gold reasoning} trajectories that can later support guideline induction.
For each supervised pair $(x^{(i)}, y^{*(i)})$, we generate candidate trajectories
$\mathbf{z} \sim P_\theta(\mathbf{z}\mid x^{(i)})$ and retain only those producing the correct answer.

\paragraph{Label-Guided Generation.}
If an initial trajectory yields an incorrect prediction, we regenerate a label-consistent explanation by conditioning on the ground-truth answer,
$\mathbf{z} \sim P_\theta(\mathbf{z} \mid x^{(i)}, y^{*(i)})$,
ensuring logical coherence aligned with supervision.

\paragraph{Structured Exploration and Memory Augmentation.}
To increase diversity, we optionally apply structured exploration (e.g., Tree-of-Thought) to generate multiple candidates.
When verified trajectories are insufficient, we retrieve similar past cases from a memory buffer $\mathcal{M}_{\text{mem}}$,
induce provisional guidelines, and generate refined trajectories conditioned on them.
Only trajectories yielding the correct answer are retained.

\paragraph{Closed-Loop Growth.}
All verified trajectories are appended to memory,
\begin{equation}
\mathcal{M}_{\text{mem}} \leftarrow \mathcal{M}_{\text{mem}} \cup \{(x^{(i)}, \mathbf{z}^{(i)*})\},
\end{equation}
forming a self-improving loop in which accumulated experience continuously expands the reasoning repository.

\subsection{Inference: Guideline Induction and Multi-Branch Reasoning}

At inference time, given a new input $x$, the model retrieves $K$ semantically similar examples from the memory repository using embedding similarity.
From their verified reasoning trajectories, it constructs a set of $M$ abstract guidelines
$\mathcal{G}_x = \{\mathbf{g}^{(1)}, \dots, \mathbf{g}^{(M)}\}$,
where each guideline represents a reusable, high-level reasoning plan induced from past successful solutions and captures shared reasoning structure rather than instance-specific details.

Guided by $\mathcal{G}_x$, reasoning proceeds in a step-wise, multi-branch manner.
At each step $t$, each guideline independently proposes a candidate reasoning step
$z_t^{(m)} \sim P_\theta(z_t \mid \hat{\mathbf{z}}_{<t}, g_t^{(m)}, x)$,
which may  via lightweight self-correction
to obtain refined step result $\tilde{z}_t^{(m)}$.
The refined candidates are then aggregated into a single unified reasoning step,
\begin{equation}
\hat{z}_t \sim P_\theta\bigl(z_t \mid \{\tilde{z}_t^{(1)}, \dots, \tilde{z}_t^{(M)}\}, x\bigr),
\end{equation}
using confidence-based scoring or voting.
The aggregated step $\hat{z}_t$ is appended to the partial trajectory and serves as shared context for the next reasoning step,
ensuring that branching improves robustness without fragmenting the reasoning process.

\paragraph{\textbf{Multi-Inference Collaboration.}}
As an optional extension, different guideline branches can be executed by different LLMs to exploit complementary reasoning behaviors.
Given the guideline set $\mathcal{G}_x$, each model $\ell$ follows one guideline during inference,
producing a step-level candidate
$\tilde{z}_t^{(\ell)} \sim P_{\theta_\ell}(z_t \mid \hat{z}_{<t}, \mathbf{g}^{(\ell)}, x)$.
These candidates are aggregated by weighted voting,
\begin{equation}
\hat{z}_t = \arg\max_z \sum_{\ell=1}^L w_\ell \cdot s_\ell(z),
\end{equation}
where $s_\ell(z)$ denotes the confidence score of model $\ell$ and $w_\ell$ its reliability weight.
This collaborative aggregation mitigates individual model bias while preserving the same step-wise inference structure.

\section{Experiments}
\subsection{Experiment Setup}
\paragraph{Dataset}
We evaluate our approach on four benchmarks: GSM8K~\cite{GSM8k} and MATH-500~\cite{math500} for mathematical reasoning, and MBPP~\cite{austin2021program} and HumanEval~\cite{chen2021evaluating} for code generation. For GSM8K, we use 500 samples for training and 1,319 for testing. For MATH-500, covering algebra, precalculus, prealgebra, counting and probability, geometry, intermediate algebra, and number theory, we randomly sample 100 training instances from each category’s training set and use the full benchmark for testing. For MBPP and HumanEval, following AFlow~\cite{zhang2024aflow}, we sample 100 and 30 training instances, with 341 and 130 reserved for testing, respectively. Valid guidelines extracted from the training samples are employed to guide model inference.

\paragraph{Baselines}
We compare against representative manually designed reasoning methods in three categories.
\textbf{Direct inference} uses input--output prompting (IO) without intermediate steps.
\textbf{Exploratory reasoning} includes CoT~\cite{cot}, CoT-SC~\cite{sc_cot}, and ReAct~\cite{react}, as well as tree-structured variants ToT~\cite{tot} and FoT~\cite{fot}.
\textbf{Search-based methods} include AFlow~\cite{zhang2024aflow}, which leverages MCTS to discover reusable reasoning workflows.

\begin{figure}[!t]
    \centering
     \includegraphics[width=0.95\linewidth]{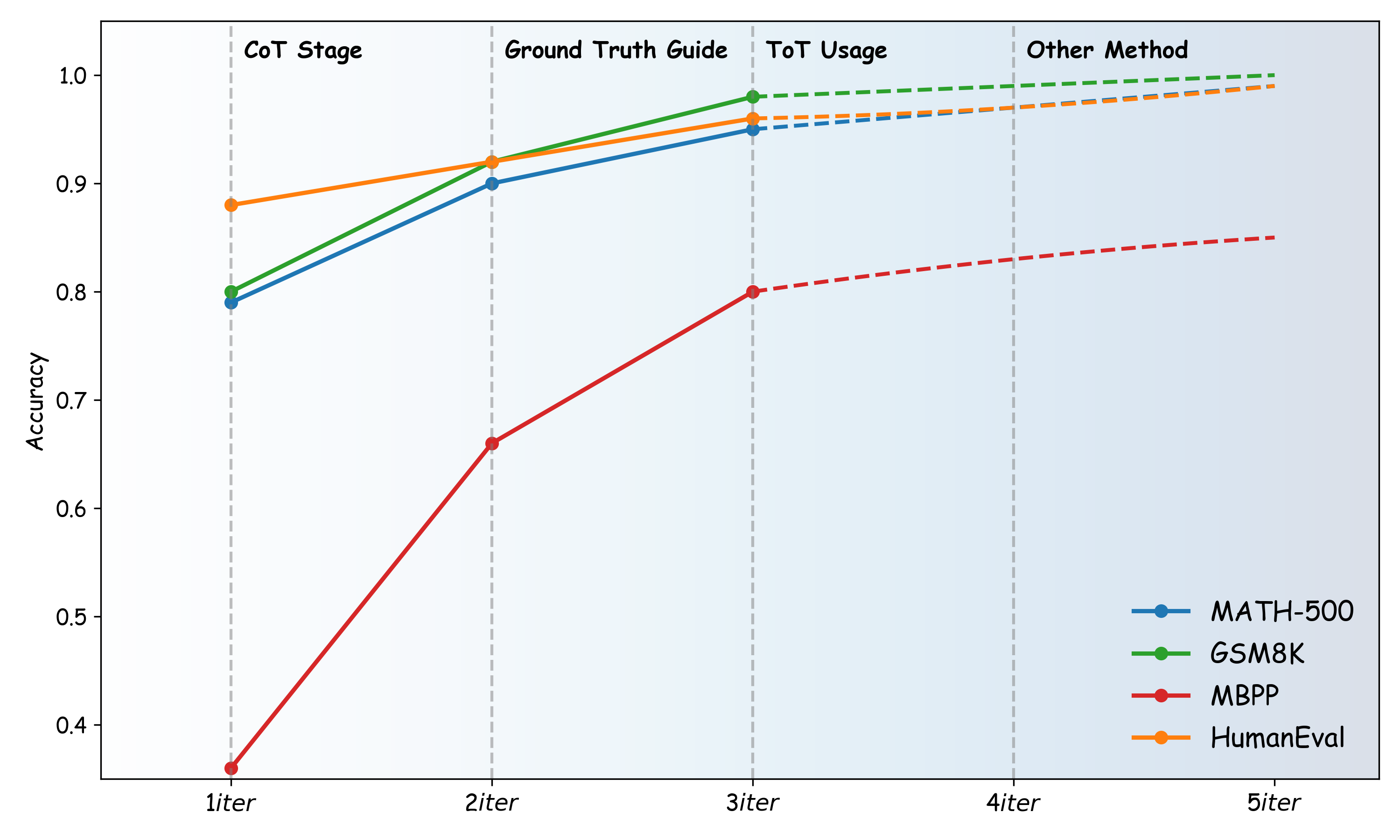}
    \caption{Illustration of the relationship between training accuracy and different stage of iterations.}
    \label{train_acc}
    \vspace{-0.5cm}
\end{figure}

\paragraph{Implementation Details}
We use GPT-4o-mini~\cite{openai2024gpt4omini} for reasoning and GPT-4o~\cite{gpt4o} for verification (Figure~\ref{fig:framework}).
\textbf{Training:} we construct gold reasoning trajectories via CoT prompting, label guidance, and ToT-based exploration; verified correct trajectories (problem, steps, embeddings) are stored, while failed cases are discarded.
Incorrect cases are refined with label guidance using a best-of-$n$ strategy (up to three rounds), and ToT~\cite{tot} is applied for harder instances; the resulting improvements are shown in Figure~\ref{train_acc}.
We compute embeddings with text-embedding-3-small.
\textbf{Inference:} we retrieve the most relevant positive sample by cosine similarity, extract high-level strategies with a Guideline Generation Prompt, rewrite them into three diverse branches, execute them step by step with per-step aggregation and refinement, and finally produce the answer with a Conclusion Prompt.
We set maximum input lengths to 400 tokens for MATH-500~\cite{math500} and GSM8K~\cite{GSM8k}, and 600 tokens for MBPP~\cite{austin2021program} and HumanEval~\cite{chen2021evaluating}.
For multi-model collaboration, we use uniform weights $w_\ell = 1$.
All prompts are provided in the Appendix.

\paragraph{Metrics} For math reasoning benchmarks (MATH-500\cite{math500} and GSM8K\cite{GSM8k}), we employ Solve Rate (\%), defined as the proportion of problems for which the model produces a correct final answer, as the primary evaluation metric. For code generation benchmarks (HumanEval\cite{chen2021evaluating} and MBPP\cite{austin2021program}), we adopt pass@1, denoting the fraction of problems for which a single generated solution successfully passes all test cases, to quantify model performance in code generation.

\begin{table*}[!t]
\small
\caption{
Comparison with current state-of-the-art methods on mathematics and code benchmarks, where
\checkmark\ denotes support and \checkmark\checkmark\ indicates strong support.
}
\label{tab:main}
\centering
\renewcommand{\arraystretch}{1.3}
\setlength{\tabcolsep}{8pt}
\begin{tabular}{lcc|cccc|c}
\toprule
\textbf{Method} & 
\makecell{\textbf{Multi-Path}}& 
\makecell{\textbf{Experience Reuse}} &
\textbf{GSM8K} & \textbf{MATH-500} & \textbf{MBPP} & \textbf{HumanEval} & \textbf{Avg.} \\
\midrule
IO              & -- & -- & 85.7  & 48.6  & 63.6 & 87.3 & 71.3 \\
CoT             & -- & -- & 87.2  & 57.8  & 66.3 & 88.6 & 75.0 \\
CoT-SC          & \checkmark & -- & 89.0  & 60.4  & 71.8 & 91.4 & 78.2 \\
ReAct           & -- & -- & 78.2  & 58.0  & 60.0 & 92.0 & 72.1 \\
ToT             & \checkmark & -- & 90.1  & 65.2  & 67.3 & 90.0 & 78.2 \\
\midrule
FoT             & \checkmark\checkmark & \checkmark (retrieval enhance) & 90.4 & 65.8 & 69.1 & 93.1 & 79.6 \\
AFlow           & -- & \checkmark\checkmark (heuristic search) & 93.1 & 65.6 & \textbf{83.4} & 94.7 & 84.1 \\
\rowcolor{gray!15}
Ours            & \checkmark\checkmark & \checkmark\checkmark (experience-induced) & \textbf{93.5} & \textbf{69.2} & 81.6 & \textbf{95.4} & \textbf{84.9} \\
\bottomrule
\end{tabular}
\end{table*}

\begin{table*}[!t]
\caption{Comparison with state-of-the-art methods on MATH-500 and HumanEval using different LLM backbones.}
\small
\label{tab:math-different-model}
\centering
\renewcommand{\arraystretch}{1.25}
\setlength{\tabcolsep}{8.4pt} 
\begin{tabular}{llcccccccc}
\toprule
\textbf{Dataset} & \textbf{Model} & IO & CoT & CoT-SC & ReAct & ToT & FoT & AFlow & \textbf{Ours} \\ 
\midrule
\multirow{3}{*}{MATH-500} 
  & GPT-4o-mini          & 48.6 & 57.8 & 60.4 & 58.0 & 65.2 & 65.8 & 65.6 & \textbf{69.2} \\
  & GPT-4o               & 51.2 & 60.4 & 63.2 & 54.8 & 67.6 & 68.4 & 66.6 & \textbf{71.4} \\
  & Qwen2.5-32B-Instruct & 50.0 & 59.0 & 61.4 & 56.4 & 66.2 & 66.6 & 64.4 & \textbf{67.6} \\ 
\midrule
\multirow{3}{*}{HumanEval} 
  & GPT-4o-mini          & 87.0 & 88.6 & 91.6 & 92.0 & 90.0 & 93.1 & 94.7 & \textbf{95.4} \\
  & GPT-4o               & 93.9 & 93.1 & 94.7 & 95.3 & 92.5 & 94.4 & 96.2 & \textbf{98.0} \\
  & Qwen2.5-32B-Instruct & 89.0 & 89.5 & 90.3 & 93.5 & 91.4 & 93.7 & 94.5 & \textbf{95.1} \\ 
\bottomrule
\end{tabular}
\end{table*}

\subsection{Comparison to State-of-the-art Methods}
\paragraph{Main Results}
We evaluate \textbf{Guideline Forest} on two categories of multi-step reasoning tasks: \textit{mathematical problem solving} (GSM8K, MATH-500) and \textit{code generation} (MBPP, HumanEval). As shown in Table~\ref{tab:main}, our method achieves strong performance across all benchmarks using GPT-4o-mini, with \textbf{93.5\%} on GSM8K and \textbf{69.2\%} on MATH-500. On HumanEval, it attains the highest accuracy of \textbf{95.4\%}. For MBPP, it achieves \textbf{81.6\%}, slightly below AFlow’s \textbf{83.4\%}, yet outperforming other approaches by a notable margin. We attribute this to AFlow’s task-specific search optimization, whereas Guideline Forest maintains higher modularity and generality across task types. These results reflect the benefit of combining two key principles: \textit{multi-path execution} and \textit{experience reuse}. Compared to methods that either sample multiple outputs or rely on fixed workflows, Guideline Forest induces reusable guidelines from verified trajectories and executes them in parallel with step-wise aggregation. This design supports stable, interpretable, and adaptable reasoning without requiring costly search. Across both symbolic and generative tasks, it offers a strong balance between accuracy, efficiency, and structural generalization.

\paragraph{Comparison with Different Base Models}
Table~\ref{tab:math-different-model} reports results across multiple LLM backbones on MATH-500 and HumanEval.
Performance consistently improves with model scale.
Under our method, GPT-4o achieves the best results on both benchmarks (71.4\% on MATH-500 and 96.2\% on HumanEval), outperforming GPT-4o-mini and Qwen2.5-32B-Instruct.
Other prompting methods such as CoT and FoT exhibit similar scaling trends.

Overall, Guideline Forest scales favorably with model capacity and consistently outperforms baseline reasoning methods.
Notably, its advantage grows with stronger models—especially on challenging tasks like MATH-500—indicating that experience-induced guidelines and step-wise aggregation better leverage the reasoning capacity of advanced LLMs.

\paragraph{Computational Cost Comparison.}
For convenience, we report inference cost on the \textbf{Num. Th.} subset of MATH-500, as summarized in Table~\ref{tab:cost-comparison}.
Guideline Forest uses 12.6k tokens per problem, compared to 7.2k for ToT, 5.4k for ReAct, and 21.8k for FoT.
Although GF requires more decoding than ReAct and ToT, it achieves substantially higher accuracy (85.48\% vs.\ 75.81\% and 67.74\%) and uses about 40\% fewer tokens than FoT while outperforming it.

To summarize inference efficiency, we adopt a simple accuracy--cost utility defined as accuracy minus a linear token penalty.
Under this metric, Guideline Forest achieves the most favorable trade-off among all methods.
Despite executing multiple branches, inference remains lightweight, as only a small number ($\sim$3) of experience-guided paths are evaluated and aggregated at each step.
As a result, inference cost scales linearly with the number of branches and reasoning steps, i.e., $\mathcal{O}(MT)$.

\begin{table}[t]
\centering
\small
\caption{Inference cost and accuracy--cost utility comparison on \textbf{Num. Th.} from MATH-500.
Acc--1kTok is defined as $\text{Acc} - \text{Tok}/1000$ with a fixed token penalty.
$T$ denotes the number of reasoning steps, $M$ the number of parallel candidates per step,
and $K$ the number of parallel trees used in FoT.}
\label{tab:cost-comparison}
\renewcommand{\arraystretch}{1.15}
\setlength{\tabcolsep}{4.7pt}
\begin{tabular}{lcccc}
\toprule
\textbf{Method} 
& \textbf{Tokens} 
& \textbf{Acc. (\%)} 
& \textbf{Acc--1kTok} 
& \textbf{Comp.} \\
\midrule
ReAct        
& \textbf{5.4k} 
& 67.74 
& 62.34 
& $\mathcal{O}(T)$ \\

ToT          
& 7.2k  
& 75.81 
& 68.61 
& $\mathcal{O}(MT)$ \\

FoT          
& 21.8k 
& 83.87 
& 62.07 
& $\mathcal{O}(KMT)$ \\

\midrule
\rowcolor{gray!15}
\textbf{Ours} 
& 12.6k 
& \textbf{85.48} 
& \textbf{72.88} 
& $\mathcal{O}(MT)$ \\
\bottomrule
\end{tabular}
\vspace{-0.2cm}
\end{table}

\begin{table*}[t]
\centering
\renewcommand{\arraystretch}{1.3}
\small
\caption{Ablation study on reasoning configurations across various mathematical subdomains.}
\label{abl}
\resizebox{\textwidth}{!}{
\begin{tabular}{l|ccccccc|c}
\toprule
\textbf{Method} & \textbf{Algebra} & \textbf{Precal.} & \textbf{Prealg.} & \textbf{Count/Prob.} & \textbf{Geom.} & \textbf{Int. Alg.} & \textbf{Num. Th.} & \textbf{Acc} \\
\midrule
w/o multi-guideline   & 84.68 & 35.71 & 73.17 & \textbf{76.32} & 41.66 & 39.18 & 79.03 & 61.39 \\
w/o aggregation        & 86.29 & 32.13 & 70.73 & \textbf{76.32} & 41.34 & 41.42 & 85.48 & 62.19 \\
w/o refinement         & 86.29 & 35.71 & 76.83 & 73.68 & 43.90 & 46.39 & 85.48 & 64.63 \\
\rowcolor{gray!15}
\textbf{Full (Ours)}   & \textbf{87.10} & \textbf{42.68} & \textbf{80.20} & 73.68 & \textbf{53.66} & \textbf{50.52} & \textbf{85.48} & \textbf{69.20} \\
\bottomrule
\end{tabular}
}
\label{tab:reasoning_ablation_compact}
\end{table*}

\subsection{Ablation Studies on Mechanism}

\begin{figure}[]
    \centering
    \begin{subfigure}[t]{0.48\linewidth} 
        \centering
        \includegraphics[width=\linewidth]{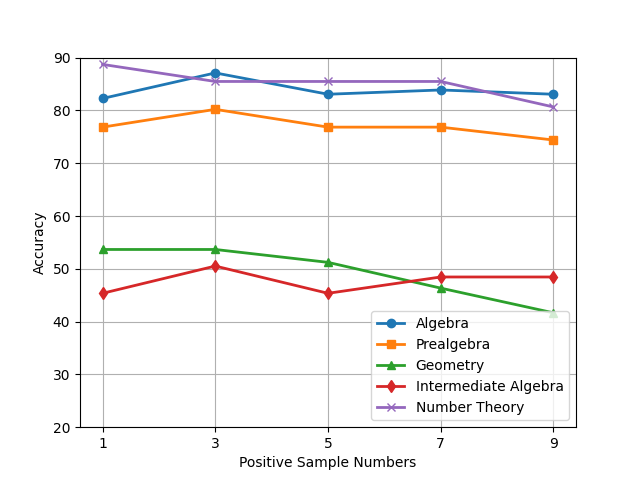}
        \caption{}
    \end{subfigure}
    \hfill
    \begin{subfigure}[t]{0.49\linewidth} 
        \centering
        \includegraphics[width=\linewidth]{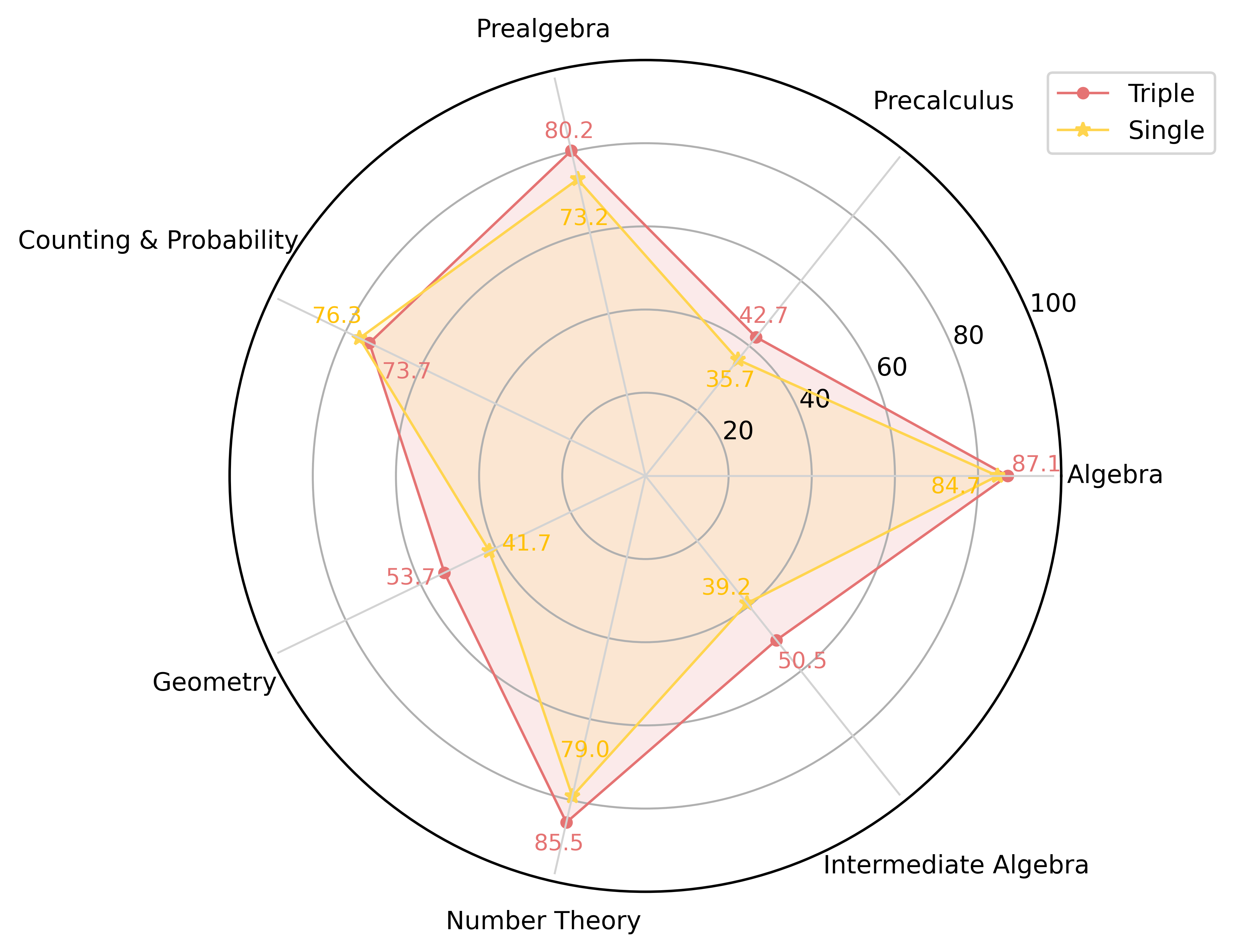}
        \caption{}
    \end{subfigure}
    \begin{subfigure}[t]{0.49\linewidth} 
        \centering
        \includegraphics[width=\linewidth]{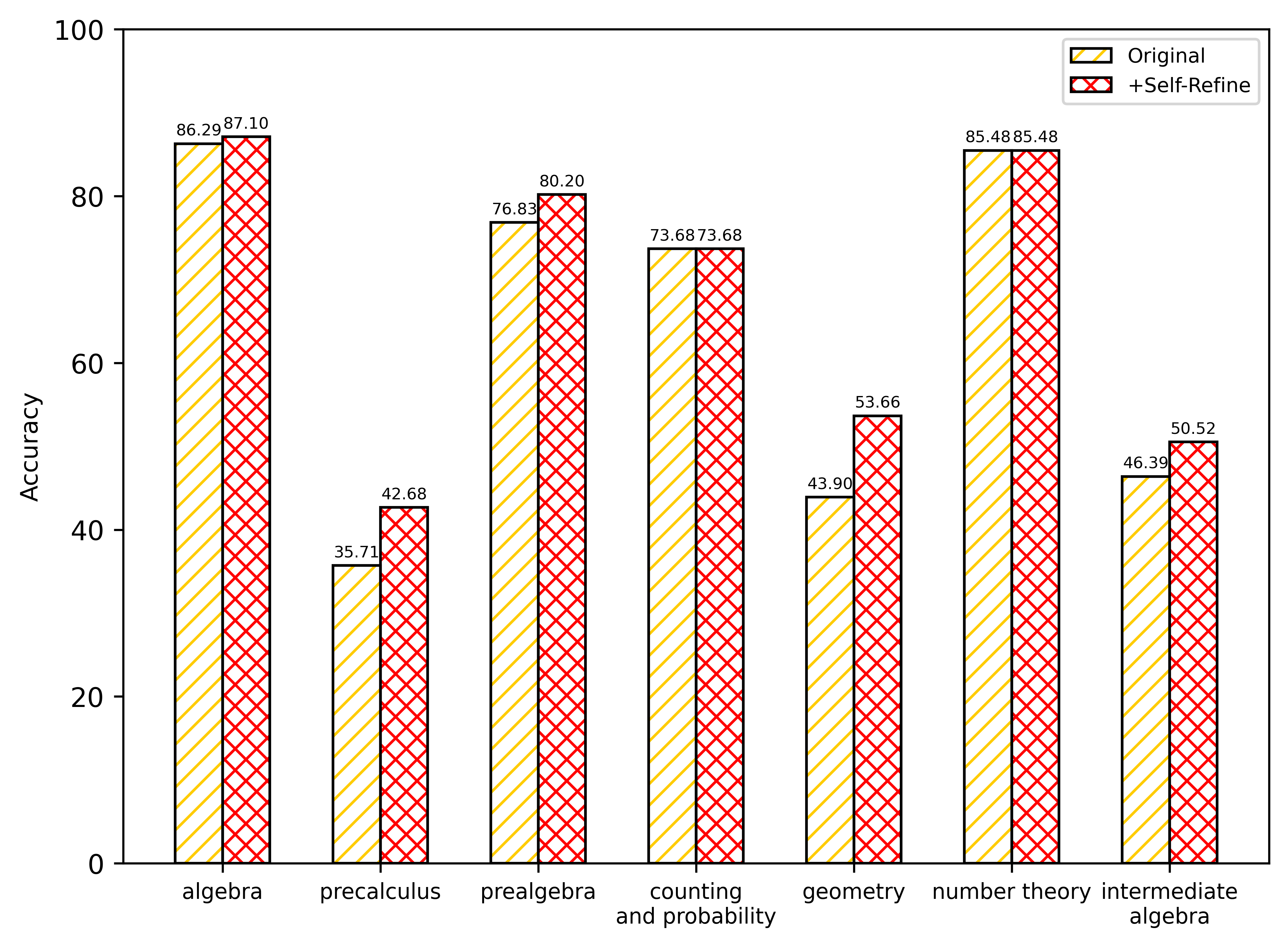}
        \caption{}
    \end{subfigure}
    \hfill
    \begin{subfigure}[t]{0.49\linewidth} 
        \centering
        \includegraphics[width=\linewidth]{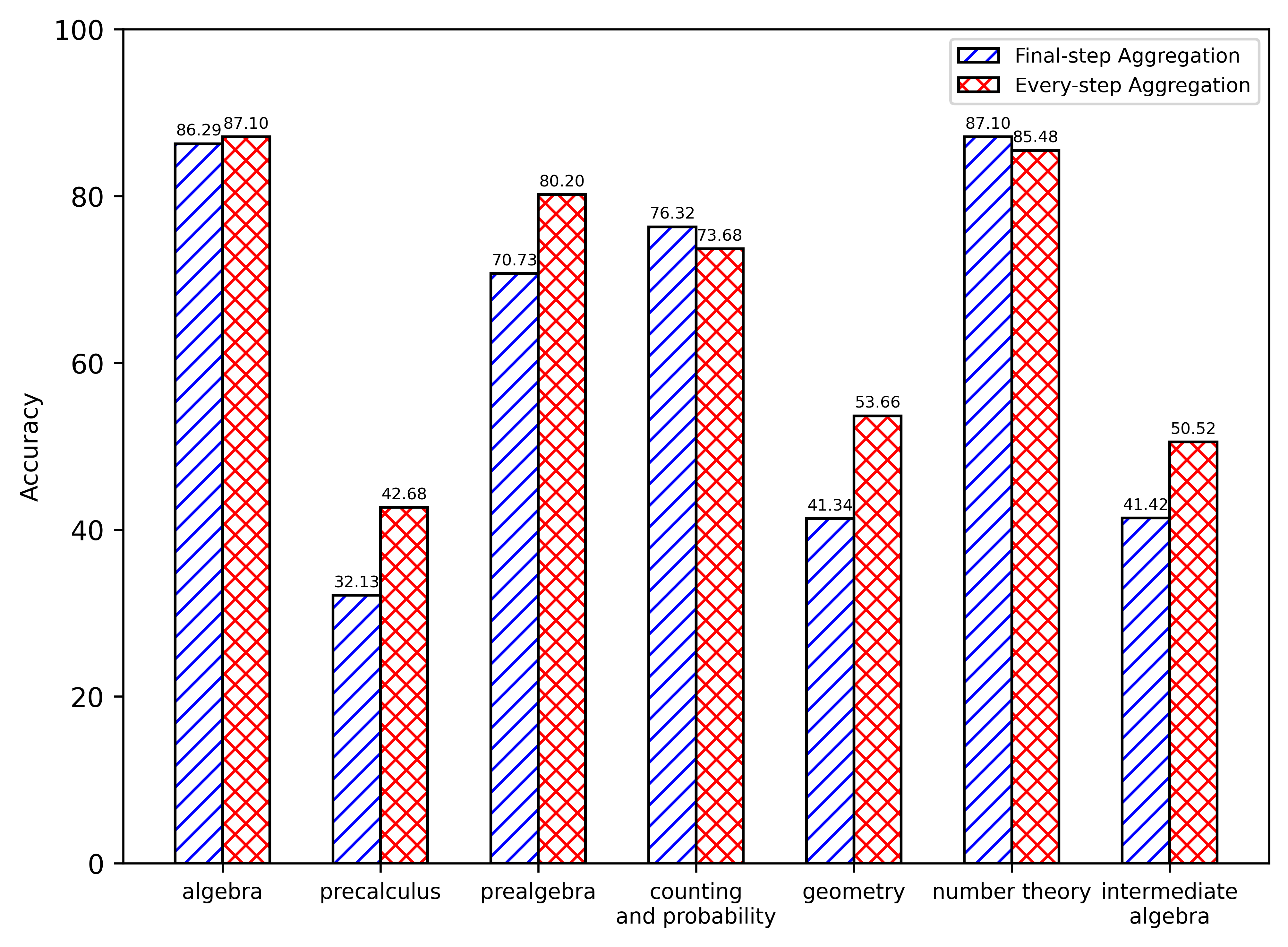}
        \caption{}
    \end{subfigure}
    \caption{Ablation studies on MATH-500 dataset evaluating the effects of (a) positive sample count, (b) reasoning path count, (c) self-refinement operation, and (d) aggregation strategies.}
    \label{Fig:1}
    \vspace{-0.2cm}
\end{figure}

\paragraph{\textbf{Retrieval Scale}} To construct a guideline for each test sample, we retrieve top-$k$ similar instances from a gold reasoning set based on cosine similarity. The solution steps of these retrieved samples are aggregated to form the guideline. Varying the value of $k$ directly impacts the quality and specificity of the resulting guideline. As shown in As shown in Figure~\ref{Fig:1}(a), aggregating three positive samples yields the best average performance. We find that too few samples produce overly specific reasoning paths that may misalign with the test instance, while too many introduce generalization noise that undermines inference precision. In contrast, three positive samples strike a favorable balance between preserving domain-specific structure and avoiding noise. It shows that fewer but more focused examples, the guideline retains more domain-specific structure, leading to stronger inductive signals.

\paragraph{\textbf{Path Diversity}}
We compare single-path reasoning with a multi-path setting, where a guideline is rewritten into multiple variants that are executed independently and aggregated.
As shown in Figure~\ref{Fig:1}(b) and Table~\ref{abl}, multi-path reasoning consistently improves accuracy across math domains (e.g., from 41.66\% to 53.66\% on Geometry).
This indicates that moderate path diversity mitigates local reasoning errors while maintaining overall consistency.
Although derived from the same guideline, diverse rewrites introduce complementary perspectives, and aggregation yields an ensemble-like robustness effect.

\begin{table}[t]
\centering
\small
\caption{Guideline-enhanced performance across different reasoning methods}
\label{tab:guideline_performance}
\renewcommand{\arraystretch}{1.1}
\setlength{\tabcolsep}{8pt}
\begin{tabular}{lcc}
\toprule
\textbf{Method} & \textbf{w/o Guideline} & \textbf{w/ Guideline} \\
\midrule
CoT   & 57.8 & 59.8 \\
ReAct & 58.0 & 66.0 \\
ToT   & 65.2 & 66.0 \\
Ours  & --   & \textbf{69.8} \\
\bottomrule
\end{tabular}
\vspace{-0.2cm}
\end{table}

\begin{figure}[t]
    \centering
    \includegraphics[width=1.0\linewidth]{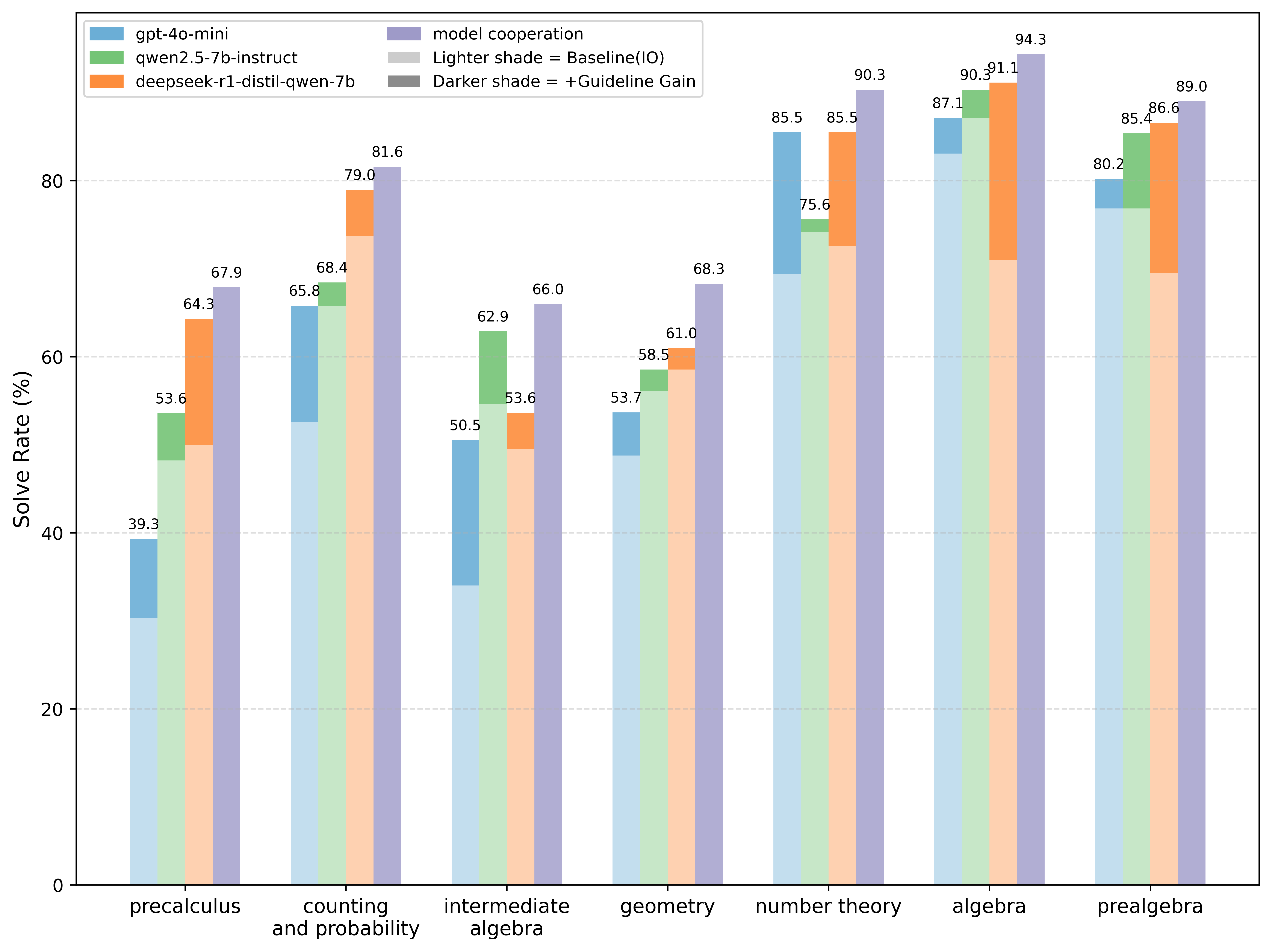}
    \caption{Comparative experiments are conducted to highlight the impact of guideline on different models and the additional benefit from model collaboration.}
    \label{fig:guideline_collaboration}
    \vspace{-0.2cm}
\end{figure}

\begin{figure*}[!ht]
    \centering
     \includegraphics[width=\linewidth]{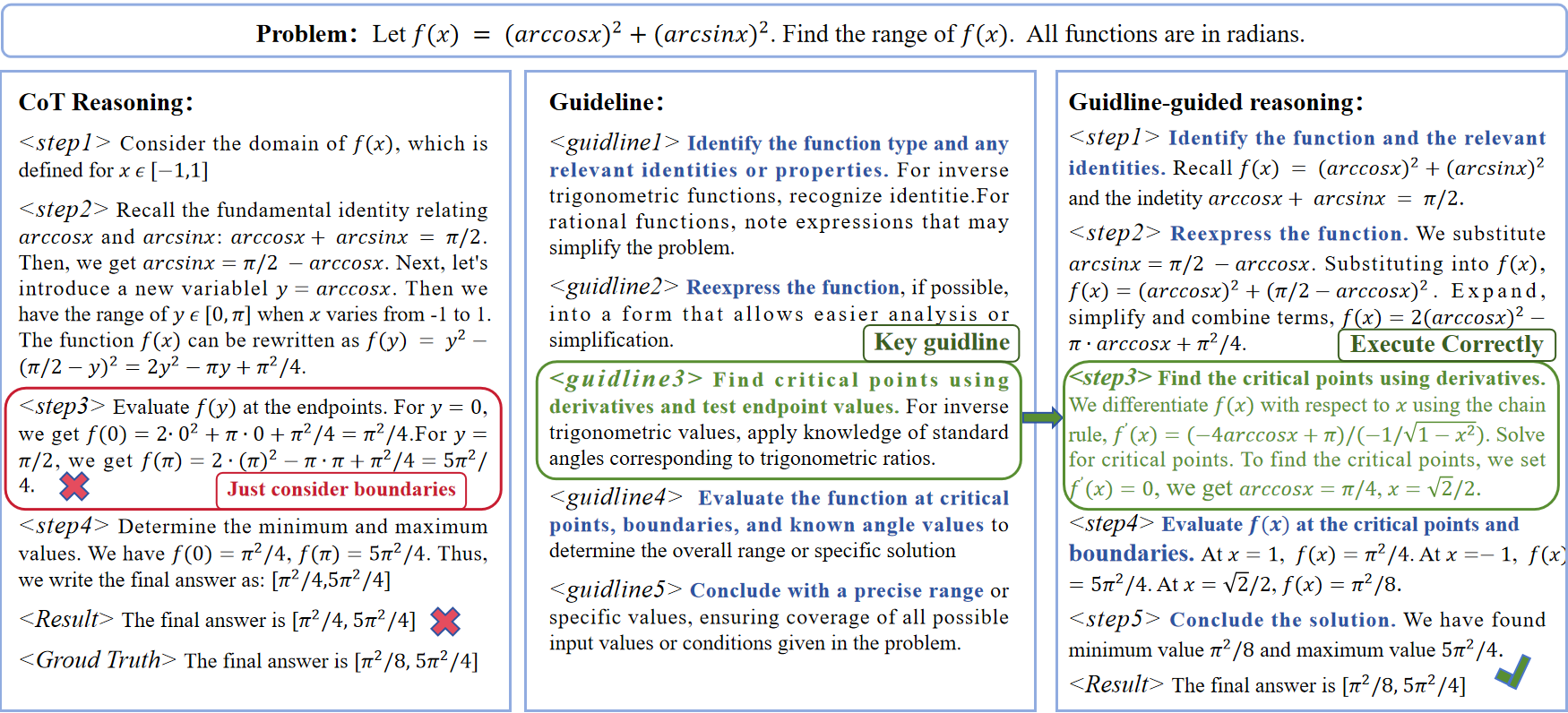}
    \caption{Qualitative experiments are conducted in mathematical scenarios and compared with the CoT method to emphasize the significant role of guideline in the reasoning process.}
    \label{Fig:case}
    \vspace{-0.2cm}
\end{figure*}

\paragraph{\textbf{Stepwise Refinement}} We introduce a step-level refinement mechanism to revise intermediate outputs before proceeding. As shown in Figure~\ref{Fig:1}(c) and Table~\ref{abl}, this improves average accuracy by about 4\% (e.g., Prealgebra from 76.83\% (w/o multi-refinement) to 80.20\% (Full)). The improvement is because refinement reduces error accumulation and acts as a local optimization loop, enhancing both the fluency and factual precision of multi-step reasoning especially in longer multi-step tasks.

\paragraph{\textbf{Aggregation Strategy}}

We compare final-step aggregation, which merges outputs only after full reasoning, with stepwise aggregation that integrates results at each step.
As shown in Figure~\ref{Fig:1}(d) and Table~\ref{abl}, stepwise aggregation consistently performs better, yielding about 7\% absolute gains. While Count/Prob. tasks remain sensitive to intermediate ambiguities, stepwise aggregation still improves overall robustness.
These results show that stepwise aggregation corrects errors early, leading to more stable and accurate reasoning.

\subsection{Extending the Role of Guidelines}

We further examine how guidelines extend beyond individual reasoning processes—enhancing diverse reasoning paradigms and facilitating collaboration among multiple models.

\paragraph{\textbf{Enhancing Different Reasoning Methods.}}
To evaluate the generality of the collected guidelines, we incorporate them into three representative reasoning paradigms—CoT, ReAct, and ToT—under identical experimental settings. 
As shown in Table 4, adding guidelines consistently improves accuracy across all methods, with the largest gains observed for ReAct and ToT. 
These results indicate that the guidelines distilled from verified reasoning trajectories are not bound to any single paradigm but capture transferable heuristics that strengthen structured and goal-directed inference. 
Such evidence highlights their utility as reusable reasoning priors that enhance diverse inference strategies.

\paragraph{\textbf{Multi-model Collaboration}}
We further evaluated the effect of the collected guidelines on different models as well as the impact of model collaboration. As shown in Figure 5, across multiple mathematical domains, all three models (GPT-4o-mini, Qwen2.5-7B-Instruct, and DeepSeek-R1-Distil-Qwen-7B) achieved consistent performance improvements when applying the guidelines. Building on this, we define model cooperation as assigning the three guidelines to the three models during the reasoning phase, where each model independently generates intermediate reasoning steps following its assigned guideline. These step-level results are then aggregated to form a more robust and coherent reasoning trajectory. Experimental results show that model cooperation outperforms single-model settings across all domains and achieves the best performance of 90.32\% on number theory tasks, demonstrating that guidelines not only enhance the reasoning ability of individual models but can also be further amplified through cooperation, thereby strengthening consistency and complementarity among models.

\subsection{Qualitative Experimental Analysis}

To illustrate how our approach influences the reasoning process, we analyze representative mathematical cases.
As shown in Figure~\ref{Fig:case}, Chain-of-Thought (CoT) reasoning often relies on boundary values when determining ranges, overlooking critical points such as local extrema and leading to incomplete or incorrect conclusions.
In contrast, our approach retrieves structurally similar examples from a verified corpus and derives an explicit reasoning strategy that emphasizes identifying critical points via derivatives.
This structured guidance clarifies the intended reasoning path, helps the model avoid common pitfalls, and results in a more principled and interpretable inference process that yields the correct solution.

Beyond this example, additional case studies in the Appendix further demonstrate that our approach consistently steers reasoning toward the correct solution space and supports iterative refinement at the step level.
Across diverse cases, the model exhibits improved robustness and accuracy by progressively optimizing intermediate reasoning steps, highlighting the effectiveness of structured experience in shaping reliable reasoning behavior.


\section{Conclusion}

We propose \textbf{Guideline Forest}, a reasoning framework that reuses verified reasoning trajectories as structured guidelines with multi-path execution and step-wise aggregation, achieving a favorable accuracy--cost trade-off with linear inference complexity.
Experiments on four math and code benchmarks show consistent gains over strong baselines, with ablations validating the roles of selective retrieval, path diversity, and aggregation.
Moreover, the framework supports collaborative inference across multiple models, enabling effective integration of complementary strengths.
Overall, our results show that grounding inference in structured reasoning experience leads to more robust and interpretable LLM reasoning.

\section*{Limitations}
Although we explore inter-model collaboration during refinement, the design space of combining models at different scales remains underexplored. Future work may investigate more diverse and adaptive model configurations to further enhance efficiency, robustness, and scalability in practical deployments.


\bibliographystyle{acl_natbib}

\clearpage

\appendix
\newpage

\section{Appendix}

\begin{tcolorbox}[
    width=\textwidth,
    colframe=black!60,
    colback=gray!5,
    fonttitle=\bfseries,
    fontupper=\footnotesize,
    title={Prompt Template: CoT Reasoning},
    breakable,
    enhanced,
    sharp corners=southwest,
    boxrule=0.5pt,
    left=2mm, right=2mm, top=1mm, bottom=1mm
]
    \textbf{Task Description:}You are a math expert skilled in step-by-step reasoning. Solve the problem methodically, clearly showing each step of your reasoning. Wrap the final answer in the format: <answer>Your Answer</answer>. Ensure your response is concise and well-structured.\\[0.5em]

\textbf{Question:} \texttt{<Your Question>}\\[0.5em]

\textbf{Solution:} \texttt{<Output Solution>}
\end{tcolorbox}

\begin{tcolorbox}[
    width=\textwidth,
    colframe=black!60,
    colback=gray!5,
    fonttitle=\bfseries,
    fontupper=\footnotesize,
    title={Prompt Template: Rectify Reasoning},
    breakable,
    enhanced,
    sharp corners=southwest,
    boxrule=0.5pt,
    left=2mm, right=2mm, top=1mm, bottom=1mm
]
    \textbf{Task Description:}You are a math expert skilled in step-by-step reasoning. Solve the problem methodically, clearly showing each step of your reasoning. Wrap the final answer in the format: <answer>Your Answer</answer>. Ensure your response is concise and well-structured.\\

    The question which you are facing is <Your Question>\\
    The correct answer is <Ground Truth>\\

    Please reference the correct answer, give the correct reasoning results.\\
    
    \textbf{Solution:} <Output Solution> 
\end{tcolorbox}

\begin{tcolorbox}[
    width=\textwidth,
    colframe=black!60,
    colback=gray!5,
    fonttitle=\bfseries,
    fontupper=\footnotesize,
    title={Prompt Template: Guideline Generation Prompt},
    breakable,
    enhanced,
    sharp corners=southwest,
    boxrule=0.5pt,
    left=2mm, right=2mm, top=1mm, bottom=1mm
]
    \textbf{Task Description:} You're an expert at math reasoning. You need to extract the core of a problem and chain of thought from examples as concise as possible. You need to complete the below tasks:

    1. Observe the Positive Samples: Carefully analyze the problem descriptions and solutions of each positive sample.\\
    2. Summarize the Guideline: Based on these samples, summarize a general guideline as concise as possible for solving this type of problem.\\
    
    \textbf{Output Format}\par
    Problem: <core problem>\par
    Chain of Thought:\par
        <step 1>: [First step of the guideline]\par
        <step 2>: [Second step of the guideline]\par
        <step 3>: [Third step of the guideline]\\
    
    Please format your summary using the above step-by-step structure. Ensure that each step is as specific and concise as possible and directly addresses the problem.
\end{tcolorbox}

\begin{tcolorbox}[
    width=\textwidth,
    colframe=black!60,
    colback=gray!5,
    fonttitle=\bfseries,
    fontupper=\footnotesize,
    title={Prompt Template: Guideline Rewriting Prompt},
    breakable,
    enhanced,
    sharp corners=southwest,
    boxrule=0.5pt,
    left=2mm, right=2mm, top=1mm, bottom=1mm
]
    \textbf{Task Description:}You are a math expert and kill at reasoning. You need to complete the below tasks:
    1.Rewrite the Guideline from Different Perspectives: Provide multiple versions of the guideline, each emphasizing a different aspect or approach.\\
    2. Maintain the Same Structure: Ensure each version has the same number of steps and maintains the logical order of the original guideline.\\
    3. Ensure Clarity and Conciseness: Make sure each version is clear and to the point.\\
    
    \textbf{Original Problem:} <Your Question>\\
    
    \textbf{Original Guideline:} <Your Chain of Thought>\\

    \textbf{Output Format:}\\
    New Guideline: \\
        <step 1>: [First step of the guideline]\\
        <step 2>: [Second step of the guideline]\\
        <step 3>: [Third step of the guideline]

\end{tcolorbox}
\clearpage
\begin{tcolorbox}[
    width=\textwidth,
    colframe=black!60,
    colback=gray!5,
    fonttitle=\bfseries,
    fontupper=\footnotesize,
    title={Prompt Template: Guideline Execution Prompt},
    breakable,
    enhanced,
    sharp corners=southwest,
    boxrule=0.5pt,
    left=2mm, right=2mm, top=1mm, bottom=1mm
]
    \textbf{Task Information:} You are a math expert. Your task is to solve a given problem by executing a guideline provided for each step to reach the correct answer.\\

    \textbf{Question:} <Your Question>\\

    \textbf{Previous analysis:} <Previous Steps>\\
    
    \textbf{Execute the guideline}: <Guideline of Current Step>. Please provide a concise result.

\end{tcolorbox}

\begin{tcolorbox}[
    width=\textwidth,
    colframe=black!60,
    colback=gray!5,
    fonttitle=\bfseries,
    fontupper=\footnotesize,
    title={Prompt Template: Aggregation Mechanism Prompt},
    breakable,
    enhanced,
    sharp corners=southwest,
    boxrule=0.5pt,
    left=2mm, right=2mm, top=1mm, bottom=1mm
]
    \textbf{Task Description:} You are an expert in reasoning and will choose the best result from the given candidates.\\

    \textbf{Question}: <Your Question>\\
    
    \textbf{Answers List}: <Answer List of Current Step>\\
    
    Select the most logically sound and informative answer. Finally, directly return <answer {} is the best>, in which {} is 1, 2, or 3. 

\end{tcolorbox}

\begin{tcolorbox}[
    width=\textwidth,
    colframe=black!60,
    colback=gray!5,
    fonttitle=\bfseries,
    fontupper=\footnotesize,
    title={Prompt Template: Refine Mechanism Prompt},
    breakable,
    enhanced,
    sharp corners=southwest,
    boxrule=0.5pt,
    left=2mm, right=2mm, top=1mm, bottom=1mm
]
    \textbf{Task Description:} You are an expert in this task and will refine the response with precise corrections. Analyze the result carefully and identify any logical errors, inconsistencies, or missing details.
    If needed, provide a refined and corrected version of the result.\\
    
    \textbf{Question:} <Your Question>\\
    
    \textbf{Current Step:} <Current Step>\\
    
    \textbf{Result:} <Your Result>

\end{tcolorbox}

\begin{tcolorbox}[
    width=\textwidth,
    colframe=black!60,
    colback=gray!5,
    fonttitle=\bfseries,
    fontupper=\footnotesize,
    title={Prompt Template: Conclusion Prompt},
    breakable,
    enhanced,
    sharp corners=southwest,
    boxrule=0.5pt,
    left=2mm, right=2mm, top=1mm, bottom=1mm
]
    \textbf{Task Information:} You are an expert in this task and please provide a structured final conclusion based ob previous analysis.\\

    \textbf{Previous Analysis}\\
    You have completed an analysis for the question: <Your Question>\\

    Here are the analysis steps performed: <Previous Steps>\\

    Based on the above analysis, summarize the final causal conclusion in a short, clear and structured manner. \\
    
    \textbf{Output format}\\
    <The final answer is: answer>

\end{tcolorbox}
\begin{tcolorbox}[
    width=\textwidth,
    colframe=black!60,
    colback=gray!5,
    fonttitle=\bfseries,
    fontupper=\footnotesize,
    title={Case Study: A Case on MATH-500 Dataset},
    breakable,
    enhanced,
    sharp corners=southwest,
    boxrule=0.5pt,
    left=2mm, right=2mm, top=1mm, bottom=1mm
]
    \textbf{Question:} Convert the point \((0,3)\) in rectangular coordinates to polar coordinates. Enter your answer in the form \((r, \theta)\), where \(r > 0\) and \(0 \leq \theta < 2\pi\). \\

    \textbf{Similarity Score:} [0.7248 0.7160 0.6174]\\

    \textbf{Positive Samples:}\\
    
    \textbf{Positive Sample 1:} \\
    \textbf{Question:} Convert the point \((-2, -2\sqrt{3}, -1)\) in rectangular coordinates to cylindrical coordinates. Enter your answer in the form \((r, \theta, z)\), where \(r > 0\) and \(0 \leq \theta < 2\pi\). \\
    \textbf{Solution:} \\
    \textbf{<step>} To convert from rectangular coordinates \((x, y, z)\) to cylindrical coordinates \((r, \theta, z)\), we use the following formulas:
    \begin{itemize}
        \item \(r = \sqrt{x^2 + y^2}\)
        \item \(\theta = \tan^{-1}\left(\frac{y}{x}\right)\)
        \item \(z\) remains the same as in rectangular coordinates.
    \end{itemize} 
\end{tcolorbox}

\clearpage

\begin{tcolorbox}[
    width=\textwidth,
    colframe=black!60,
    colback=gray!5,
    fonttitle=\bfseries,
    fontupper=\footnotesize,
    title={Case Study: A Case on MATH-500 Dataset},
    breakable,
    enhanced,
    sharp corners=southwest,
    boxrule=0.5pt,
    left=2mm, right=2mm, top=1mm, bottom=1mm
]   
    \textbf{<step>} Given the point \((-2, -2\sqrt{3}, -1)\), we identify values for \(x\), \(y\), and \(z\):
    \begin{itemize}
        \item \(x = -2\)
        \item \(y = -2\sqrt{3}\)
        \item \(z = -1\)
    \end{itemize}
    
    \textbf{<step>} First, we calculate \(r\):
    \[
    r = \sqrt{x^2 + y^2} = \sqrt{(-2)^2 + (-2\sqrt{3})^2} = \sqrt{4 + 12} = \sqrt{16} = 4.
    \] 
    \textbf{<step>} Next, we find \(\theta\):
    \[
    \theta = \tan^{-1}\left(\frac{y}{x}\right) = \tan^{-1}\left(\frac{-2\sqrt{3}}{-2}\right) = \tan^{-1}(\sqrt{3}).
    \]
    Since \(\tan^{-1}(\sqrt{3})\) corresponds to an angle of \(\frac{\pi}{3}\), we need to determine the correct quadrant.
    
    Given both \(x\) and \(y\) are negative, the point is in the third quadrant. Therefore, we add \(\pi\) to \(\frac{\pi}{3}\):\\
        \[
    \theta = \frac{\pi}{3} + \pi = \frac{\pi}{3} + \frac{3\pi}{3} = \frac{4\pi}{3}.
    \] 
    
    \textbf{<step>} Lastly, \(z\) stays the same:
    \[
    z = -1.
    \]
    
    \textbf{<step>} Thus, the cylindrical coordinates are:
    \[
    (r, \theta, z) = (4, \frac{4\pi}{3}, -1).
    \]
    
    \textbf{<The final answer is: \((4, \frac{4\pi}{3}, -1)\)>} \\
    \textbf{Positive Sample 2:} ... \\
    \textbf{Positive Sample 3:} ... \\
        \textbf{Guideline:}\\
    
    \textbf{Core Problem:} The core problem is converting a point from one coordinate system to another (rectangular to cylindrical, rectangular to spherical, or adjusting spherical coordinates for standard representation) \\

    \textbf{Reasoning Steps:}\\
    \textbf{<step 1>} Identify the given point's current coordinate system and values.\\
    \textbf{<step 2>} Apply the appropriate formulas for conversion:
    \begin{itemize}
        \item For cylindrical: \( r = \sqrt{x^2 + y^2} \), \(\theta = \tan^{-1}(y/x)\), and \(z\) remains unchanged.
        \item For spherical: \(\rho = \sqrt{x^2 + y^2 + z^2}\), \(\theta = \tan^{-1}(y/x)\), \(\phi = \cos^{-1}(z/\rho)\).
    \end{itemize}
    \textbf{<step 3>} Ensure all angles are in their respective domains:
    \begin{itemize}
        \item \(0 \leq \theta < 2\pi\)
        \item For \(\phi\) in spherical coordinates: \( 0 \leq \phi \leq \pi \).
    \end{itemize}
    \textbf{<step 4>} Adjust angles if necessary:
    \begin{itemize}
        \item Correct \(\theta\) based on quadrant, or adjust \(\phi\) if beyond standard range.\\
    \end{itemize}

    \textbf{Rewritten Guideline:}\\

    \textbf{Core Problem:} The core problem is converting a point from one coordinate system to another (rectangular to cylindrical, rectangular to spherical, or adjusting spherical coordinates for standard representation)\\

    \textbf{Reasoning Steps:}\\
    \textbf{<Step 1>} Determine the coordinate system of the provided point and note down the exact values.\\
    \textbf{<Step 2>} Execute the necessary mathematical transformations:
    \begin{itemize}
    \item Cylindrical conversion: Compute \( r = \sqrt{x^2 + y^2} \), ascertain \(\theta = \tan^{-1}(y/x)\), and retain \(z\).
    \item Spherical conversion: Calculate \(\rho = \sqrt{x^2 + y^2 + z^2}\), ascertain \(\theta = \tan^{-1}(y/x)\), and determine \(\phi = \cos^{-1}(z/\rho)\).
    \end{itemize}
    
    \textbf{<Step 3>} Confirm that computed angles fall within standard mathematical intervals:
    \begin{itemize}
        \item Ensure \(0 \leq \theta < 2\pi\).
        \item For \(\phi\) in spherical coordinates: validate \( 0 \leq \phi \leq \pi \).
    \end{itemize}
    
    \textbf{<Step 4>} Modify angles when necessary by assessing quadrant positioning or correcting \(\phi\) to maintain standard practice.
\end{tcolorbox}

\clearpage

    \begin{tcolorbox}[
    width=\textwidth,
    colframe=black!60,
    colback=gray!5,
    fonttitle=\bfseries,
    fontupper=\footnotesize,
    title={Case Study: A Case on MATH-500 Dataset},
    breakable,
    enhanced,
    sharp corners=southwest,
    boxrule=0.5pt,
    left=2mm, right=2mm, top=1mm, bottom=1mm
]
    \textbf{Execution Result:}\\

    \textbf{Step1 Result:} The current coordinate system is the rectangular (Cartesian) coordinate system, and the provided values are the coordinates of the point: \((0, 3)\).\\

    \textbf{Refined Result:} The given result correctly identifies the rectangular coordinates as \((x, y) = (0, 3)\). This statement accurately represents the initial step in the problem-solving process, where the point's coordinates are extracted for conversion. There are no logical errors or inconsistencies in this initial step. \\
    \textbf{Step2 Result:} To convert the point \((0,3)\) from rectangular coordinates to polar coordinates, we'll use the formulas for polar coordinate conversion.
    \begin{itemize}
        \item Calculate the radial distance \(r\):
        \[r = \sqrt{x^2 + y^2} = \sqrt{0^2 + 3^2} = \sqrt{9} = 3\]
        
        \item Calculate the azimuthal angle \(\theta\):
        Since \(x = 0\), to find \(\theta\), we observe the position. The point is along the positive \(y\)-axis.
        \[\theta = \frac{\pi}{2}\]
    \end{itemize}
    Thus, the polar coordinates of the point \((0,3)\) are \((3, \frac{\pi}{2})\). \\
        \textbf{Refined Result:} The original result is correct, and there are no logical errors or inconsistencies in the steps or the solution. The provided conclusion accurately represents the polar coordinates of the given rectangular coordinates.\\

    \textbf{Step3 Result:} To convert the point \((0,3)\) from rectangular to polar coordinates, we follow these steps:

    \begin{itemize}
        \item \textbf{Calculate \(r\):} 
        \[r = \sqrt{x^2 + y^2} = \sqrt{0^2 + 3^2} = \sqrt{9} = 3\]
        
        \item \textbf{Calculate \(\theta\):} 
        Since the point \((0,3)\) lies on the positive \(y\)-axis, \(\theta\) is \(\frac{\pi}{2}\).
        
        Adjust \(\theta\) to ensure \(0 \leq \theta < 2\pi\). In this case, \(\theta = \frac{\pi}{2}\) is already within the required range.
    \end{itemize}

    Therefore, the polar coordinates are \((r, \theta) = (3, \frac{\pi}{2})\).\\

    \textbf{Refined Result:} The given analysis is mostly correct, but let's ensure each step is clearly correct and complete.(Verification process omitted due to space constraints) . Thus, the polar coordinates of the point \((0, 3)\) are \((r, \theta) = (3, \frac{\pi}{2})\).
    The original solution correctly calculates the polar coordinates without errors or missing details.\\

    \textbf{Step4 Result:} The point \((0, 3)\) in rectangular coordinates can be converted to polar coordinates by calculating: \\
    \begin{itemize}
        \item \textbf{r (radius):}  
        \[r = \sqrt{x^2 + y^2} = \sqrt{0^2 + 3^2} = \sqrt{9} = 3\]
    
        \item \textbf{$\theta$ (angle):}  
        Since the point lies on the positive \(y\)-axis, \(\theta\) is \(\frac{\pi}{2}\).
    \end{itemize}
    
    Therefore, the polar coordinates are \((3, \frac{\pi}{2})\).\\

    \textbf{Refined Result:} The conversion from rectangular coordinates \((0,3)\) to polar coordinates is mostly correct, but a deeper analysis of the approach should be conducted to ensure accuracy for edge cases like this one.\\

    \textbf{Final Result:} The final answer is: The polar coordinates of the point \((0, 3)\) are \((3, \frac{\pi}{2})\).
\end{tcolorbox}

\end{document}